\documentclass{article}

\usepackage{microtype}
\usepackage{graphicx}
\usepackage{subcaption}
\usepackage{booktabs}
\usepackage[table]{xcolor}

\usepackage{hyperref}

\usepackage[preprint]{icml2026}

\usepackage{amsmath}
\usepackage{amssymb}
\usepackage{mathtools}
\usepackage{amsthm}
\usepackage[capitalize,noabbrev]{cleveref}
\usepackage{multirow}
\usepackage{makecell}
\usepackage{xspace}
\newcommand{\method}{\textsc{Grad2Reward}\xspace}
\usepackage[breakable]{tcolorbox}
\tcbset{
  promptbox/.style={
    colback=gray!5!white,        
    colframe=black!50,           
    colbacktitle=gray!10,        
    coltitle=black,              
    fonttitle=\bfseries,
    boxrule=0.4pt,
    arc=1mm,
    left=1mm,
    right=1mm,
    top=1mm,
    bottom=1mm,
    breakable,
    title={},
  }
}
\usepackage{enumitem}

\theoremstyle{plain}

\theoremstyle{definition}

\theoremstyle{remark}

\usepackage[textsize=tiny]{todonotes}

\usepackage{xcolor}

\begin{document}

\twocolumn[
  \icmltitle{Grad2Reward: From Sparse Judgment to Dense Rewards for Improving Open-Ended LLM Reasoning}



  \icmlsetsymbol{equal}{*}

  \begin{icmlauthorlist}
    \icmlauthor{Zheng Zhang}{shanghaitech,bigai}
    \icmlauthor{Ao Lu}{shanghaitech}
    \icmlauthor{Yuanhao Zeng}{shanghaitech,bigai}
    \icmlauthor{Ziwei Shan}{shanghaitech}
    \icmlauthor{Jinjin Guo}{jingdong}
    \icmlauthor{Lufei Li}{shanghaitech}
    \icmlauthor{Yexin Li}{bigai}
    \icmlauthor{Kan Ren}{shanghaitech}
  \end{icmlauthorlist}

  \icmlaffiliation{shanghaitech}{School of Information Science and Technology, ShanghaiTech University}
  \icmlaffiliation{bigai}{State Key Laboratory of General Artificial Intelligence, BIGAI}
  \icmlaffiliation{jingdong}
  {JD.com. Email: Zheng Zhang \textless zhengzhang2024@shanghaitech.edu.cn\textgreater}
    
  \icmlcorrespondingauthor{Yexin Li}{liyexin@bigai.ai}
  \icmlcorrespondingauthor{Kan Ren}{renkan@shanghaitech.edu.cn}

  \icmlkeywords{Machine Learning, ICML}

  \vskip 0.3in
]



\printAffiliationsAndNotice{}  

\begin{abstract}
Reinforcement Learning with Verifiable Rewards (RLVR) has catalyzed significant breakthroughs in complex LLM reasoning within verifiable domains, such as mathematics and programming.
Recent efforts have sought to extend this paradigm to open-ended tasks by employing LLMs-as-a-Judge to provide sequence-level rewards for policy optimization.
However, these rewards are inherently sparse, failing to provide the fine-grained supervision necessary for generating complex, long-form trajectories.
Furthermore, current work treats the Judge as a black-box oracle, discarding the rich intermediate feedback signals encoded in it.
To address these limitations, we introduce \method, a novel framework that extracts dense process rewards directly from the Judge's model inference process via a single backward pass.
By leveraging gradient-based attribution, \method enables precise \emph{token-level credit assignment}, substantially enhancing training efficiency and reasoning quality.
Additionally, \method introduces a \emph{self-judging mechanism}, allowing the policy to improve through its own evaluative signals without training specialized reward models or reliance on superior external Judges.
The experiments demonstrate that policies optimized with \method achieve outstanding performance across diverse open-ended tasks, affirming its effectiveness and broad generalizability.

\end{abstract}

\section{Introduction}
Large language models (LLMs) have recently demonstrated remarkable progress in complex reasoning tasks such as mathematics and programming, as shown by OpenAI's o-series \citep{jaech2024openai} and DeepSeek-R1 \citep{guo2025deepseek}.
A key driver of these advances is Reinforcement Learning with Verifiable Rewards (RLVR), which assigns outcome rewards by verifying policy\footnote{We use \textit{policy} to represent the underlying LLM generator for the given task query.} outputs against ground-truth answers.
RLVR relies on ground-truth labels and therefore cannot be directly applied to broader open-ended tasks \citep{xu2025direct, ye2025self}, such as medical consultation or creative writing, where evaluation is inherently subjective and not strictly verifiable.
Recent work \citep{gunjal2025rubrics, zhou2025breaking} attempts to extend RLVR to open-ended tasks by leveraging LLM-as-a-Judge to provide sequence-level rewards for policy optimization.
For example, \citet{gunjal2025rubrics} employs an LLM-as-a-judge paradigm to evaluate policy outputs along multiple criteria and convert the evaluations into sequence-level rewards for GRPO \citep{shao2024deepseekmath} optimization.

However, these methods face several limitations.
(i) \emph{Sparse rewards result in a lack of fine-grained supervision.}
Sparse rewards only provide feedback at the end of the trajectory, causing different parts of the generation sequence to be treated equally.
Unlike verifiable tasks that only care about the correctness of the final answer, open-ended tasks place greater emphasis on the quality of the process. 
For example, in medical counseling, every part of the response must be scientifically sound and non-misleading. 
This property makes open-ended tasks fundamentally require dense rewards for fine-grained optimization.
(ii) \emph{Insufficient use of Judge feedback.}
In practice, the Judge evaluates the quality and validity of the policy outputs holistically and emits a final decision.
While such a sequence-level signal reflects an overall assessment of the policy outputs, it is informed by the quality of different parts of the generation, suggesting the potential to provide guidance beyond coarse sequence-level supervision.
Despite this, prior approaches \citep{zhou2025breaking, gunjal2025rubrics, shao2025dr, bi2025reward} regard the Judge as a black box, relying solely on the final verdict as a reward, without exploiting the additional evaluative information implicit in the Judge's assessment of the full generation from the policy.
As a result, fine-grained rewards are critical but challenging to obtain, while the rich feedback signals inherent in the Judge remain underutilized.

To address these limitations, we introduce \method, a framework designed to unlock the dense feedback hidden within the Judge.
Specifically, \method leverages gradient-based attribution to quantify each generated token's contribution to the Judge's decision and converts these contributions into dense token-level rewards. 
The entire procedure consumes only a single backward pass and does not require any fine-tuning of the Judge.
These rewards are used to optimize the policy, ultimately enhancing the LLM's performance on open-ended tasks.
Notably, unlike prior work \citep{bi2025reward, zhou2025breaking, gunjal2025rubrics, wang2025infimed} that relies on more advanced models as Judge to supervise policy training, we adopt a self-judging mechanism where the Judge is fixed to the initial policy model, ensuring that policy improvement arises from its own potential rather than distilling external knowledge.

Empirically, optimizing the policy with the dense rewards provided by \method leads to superior training efficiency and better performance.
By providing dense, informative supervision, our \method significantly accelerates convergence, requiring substantially fewer training steps to reach optimal performance compared to sparse-reward baselines. 
In addition, extensive experiments confirm that \method consistently achieves strong performance across multiple domains, validating both its effectiveness and generalization capability.
Our contributions can be summarized as follows:
\vspace{-2mm}
\begin{itemize}[leftmargin=3mm]
    \item To the best of our knowledge, we are the first to introduce a dense-reward framework for improving LLM performance on open-ended tasks, effectively addressing the credit assignment problem and substantially improving training efficiency.
    \vspace{-1mm}
    \item We introduce a self-judging mechanism that allows a policy to leverage its own evaluative feedback for improvement, eliminating the need for stronger and expensive external Judges.
    \vspace{-1mm}
    \item \method achieves leading performance across multiple open-ended tasks compared to sparse-reward baselines, demonstrating strong effectiveness and broad applicability. Notably, it provides process rewards without training a dedicated process reward model (PRM) and generalizes to verifiable domains, where it shows clear advantages over PRM-based methods.
\end{itemize}

\vspace{-1.5em}

\section{Related Work}
\label{sec:related_work}

\paragraph{LLM for Open-ended Tasks.}
Unlike verifiable tasks which have well-defined ground truth, open-ended tasks such as medical consultation \citep{arora2025healthbench} or scientific question answering \citep{yifei2025researchqa} lack standard answers and are subjective, typically requiring human experts for evaluation.
Recent work \citep{zhou2025breaking, gunjal2025rubrics, shao2025dr, bi2025reward, huang2025reinforcement} has explored using LLM-as-a-Judge \citep{chen2025judgelrm, lee2025checkeval, wu2025meta} to generate feedback signals as sequence-level rewards for optimizing LLM performance on open-ended tasks through reinforcement learning (RL).
For example, \citep{gunjal2025rubrics} uses an LLM-as-a-Judge to evaluate whether the policy output satisfies specified rubrics and converts the evaluation into rewards for RL optimization.
\citep{wang2025infimed} integrates retrieval-augmented in-context prompting into an RL training framework to improve performance in medical dialogue.
However, the rewards used in these methods are sparse and therefore cannot provide fine-grained supervision over the policy’s generated sequence,
a limitation that is particularly pronounced in open-ended tasks.

\paragraph{Dense Rewards Modeling.}
Dense rewards have been shown to be effective for improving the reasoning capabilities of LLMs \citep{lightman2024lets, wang2024math}.
Recent work \citep{wang-etal-2024-math, zeng2025versaprm, li2025process} has explored training PRMs in mathematical domains, which provide step-level rewards \citep{zhang2025linking, cheng2025stop} or token-level rewards \citep{cui2025process} for policy optimization. 
For example, 
PQM \citep{li2025process} models the process reward as a Q-value ranking problem in a Markov Decision Process.
PURE \citep{cheng2025stop} first trains a PRM, and then applies it to the proposed min-form credit assignment.
VersaPRM \citep{zeng2025versaprm} applies PRM to other verifiable domains beyond mathematics.
However, these methods require ground-truth process labels to train a domain-specific reward model, making them applicable only to verifiable domains and hard to extend to open-ended tasks.

\begin{figure*}[t]
  \centering
  \includegraphics[width=\textwidth]{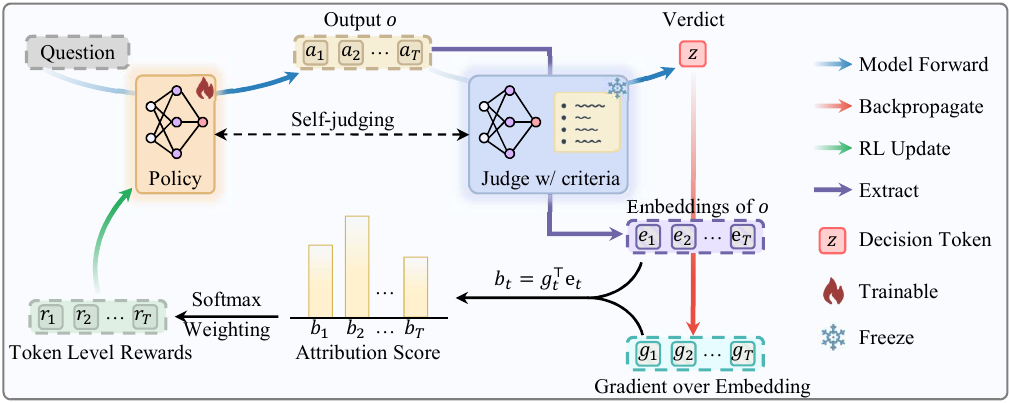}
  \caption{\textbf{Overview of \method} (better viewed in color): The Policy samples an output $o$, which is evaluated by a Judge to derive a verdict $z$. By computing the inner product between the output embeddings $e_t$ and their gradients $g_t$ (derived from $z$), we obtain token-wise attribution scores $b_t$. These are converted into token-level rewards $r_t$ to guide the policy optimization.}
  \label{fig:overview}
\end{figure*}

\section{Preliminary}
\subsection{Open-ended LLM Reasoning}
Given a query $x$, the LLM $\pi$ generates a response $o = (a_1, a_2, \dots, a_T)$, where $a_t$ denotes the $t$-th token and $T$ is the total number of tokens.
We model the generation process of the LLM as a finite-horizon Markov Decision Process (MDP) $\mathcal{M} = (\mathcal{S}, \mathcal{A}, \mathcal{P}, r)$, in which an autoregressive LLM $\pi$ serves as the policy. 
Here, $\mathcal{S}$ is the state space, $\mathcal{A}$ is the action space, $\mathcal{P}$ denotes the transition dynamics, and $r: \mathcal{S} \times \mathcal{A} \to \mathbb{R}$ is the reward function. 
For the $t$-th token, the state is defined as $s_t = (x, a_{\le t-1})$ which contains the query $x$ and the sequence of previously generated tokens $a_{\le t-1}$.
The action $a_t$ is the token generated conditioned on state $s_t$. The state transition is deterministic, as the next state is uniquely determined by concatenating the previously generated sequence with the current token $a_t$. 
Let $r_t$ denote the immediate reward of token $a_t$, such that the sequence-level reward of the policy output $o$ can be written as $r(x, o) = \Sigma_{t=1}^{T} r_{t}$. 

In open-ended tasks, prior approaches obtain sequence-level reward $r(x, o)$ by using the LLM-as-a-Judge, while the individual token rewards $r_{t}$ remain unknown.

\subsection{LLM-as-a-Judge}
\label{pre:LLM-as-a-Judge}
To evaluate a policy output $o$ for a given query $x$, existing methods \citep{zhou2025breaking, gunjal2025rubrics, shao2025dr, bi2025reward} employ the LLM-as-a-Judge combined with a predefined rubric $(c, w)$, where each $c$ specifies an evaluation criterion and $w$ is the reward assigned when the criterion is satisfied.

Specifically, a structured prompt (see Appendix~\ref{appendix:Judge_prompt}) is constructed, which includes the query $x$, the policy output $o$, and the criterion $c$ as the input to the Judge.
The Judge is instructed to generate a binary decision token $z \in \{\texttt{True}, \texttt{False}\}$ to indicate whether the criterion $c$ is satisfied.
Then, the sequence-level reward for $(x,o)$ is defined using the Judge's output as:
\begin{equation}
r(x,o|c, w)
= w \cdot \mathbb{I}[z \sim p_{\text{judge}}(\,\cdot \mid x, o, c)]
\label{eq:outcome_reward}
\end{equation}
Here, $\mathbb{I}[\cdot]$ denotes the indicator function, which evaluates to $1$ if the Judge outputs $z=\texttt{True}$, and $0$ otherwise.
The resulting sequence-level reward $r(x,o|c,w)$ is then used in RL optimization algorithms, such as GRPO, to improve the policy’s performance.
For brevity, we denote $r(x,o \mid c,w)$ as $r(x,o)$ in the following.

\section{Methodology}

\subsection{Judge Implicitly Contains Process Feedback}
Although the Judge ultimately evaluates the policy output $o$ with a binary decision token, its internal computation is far more expressive.
As an autoregressive language model, the Judge processes the entire policy output $o$ token by token before emitting a final judgment, which reflects an accumulated assessment of the whole sequence $o$ based on a semantic understanding of the intermediate steps.
During this process, Judge implicitly evaluates the generation trajectory by modeling logical structure, semantic coherence, and alignment with the given criterion.
However, existing approaches \citep{zhou2025breaking, gunjal2025rubrics, shao2025dr} treat the Judge as a black box and only extract the final verdict as a scalar reward.
This collapses the Judge’s rich internal evaluative process into a single binary signal, implicitly assigning equal credit to all tokens in the trajectory.
As a result, a large amount of intermediate feedback information that may indicate the quality of reasoning processes is discarded.
This motivates our design to recover and exploit the process feedback that is implicitly embedded in the Judge’s autoregressive computation but overlooked by existing black-box usage approaches.

\paragraph{\textbf{Self-Judging Mechanism.}} 
In this work, we adopt a self-judging mechanism in which the Judge is instantiated as a frozen copy of the initial policy rather than a stronger, more expensive external model.  
During training, the Judge remains fixed and is used solely to provide evaluation signals, while the policy is optimized via RL.
This design is motivated by the observation that \emph{LLMs often exhibit stronger discriminative capabilities than generative capabilities}, as further supported by \citet{song2025mind}.  
Freezing the Judge ensures that the feedback signal remains stable throughout training, providing consistent supervision.  
As a result, the policy improves by leveraging its own evaluative capacity rather than by distilling knowledge from a superior external Judge.  
This mechanism can thus be viewed as a form of self-improvement, where the model iteratively refines its generation behavior based on its own discriminative signals.

\subsection{Fine-Grained Reward Design}
\label{sec:via_gradient_attribution}
To extract fine-grained signals from the Judge, we employ gradient-based attribution to measure how each token contributes to the generation of the decision token $z$.
Concretely, let $\mathbf{e}_t \in \mathbb{R}^d$ denote the embedding of the $t$-th token in the policy output $o = (a_1, \dots, a_T)$.
We compute the gradient of the log-probability that the Judge generates the specific decision token $z$ with respect to each token embedding:
\begin{equation}
\mathbf{g}_{t}
=
\nabla_{\mathbf{e}_t}
\log p_{\text{judge}}(z \mid x, o, c)
\label{eq:token_gradient}
\end{equation}

Gradient vectors $\mathbf{g}_{t}$ are next transformed into scalar attribution scores that reflect the importance of each token for the Judge's decision using \emph{Gradient $\times$ Embedding}.
\begin{equation}
b_{t}
=
\mathbf{g}_{t}^{\top} \mathbf{e}_t
\label{eq:gradient_attribution}
\end{equation}

Eq.~\eqref{eq:token_gradient} yields a sequence of attribution scores
$\mathbf{b} = (b_{1}, \dots, b_{T})$, where each $b_t$ reflects the contribution of the $t$-th token in policy output $o$ to the Judge’s decision. As the raw attribution scores $\mathbf{b} = (b_{1}, \dots, b_{T})$ can vary significantly in scale, we apply the softmax function to normalize these signals and quantify the relative contribution of each token to the Judge's decision:
\begin{equation}
\alpha_t = \text{Softmax}(\mathbf{b})_t = \frac{\exp(b_t / \tau)}{\sum_{k=1}^{T} \exp(b_k / \tau)}
\end{equation}
where $\tau$ is a temperature parameter controlling the sharpness of the distribution.
Based on these attribution scores, we decompose the sequence-level reward $r(x,o)$ into token-level rewards:
\begin{equation}
r_t = \alpha_t \cdot r(x,o)
\label{eq:token_reward}
\end{equation}
As a result, the originally sparse sequence-level supervision is transformed into dense token-level feedback, enabling more precise credit assignment through the gradient signals implicitly encoded in the Judge’s internal computation.

\subsection{Theoretical Analysis of the Reward Design}
\label{sec:theoretical_analysis}

In this subsection, we provide a theoretical perspective on the reward design.
Let the objective function represent the Judge's log-probability of generating the decision token for criterion $c$ given input $x$:
\begin{equation}
F(\mathbf{e}_1, \dots, \mathbf{e}_T) 
=
\log p_{\text{judge}}(z \mid x, o, c)
\end{equation}
Consider a local perturbation $\Delta \mathbf{e}_t$ applied independently to each token embedding $\mathbf{e}_t$.  
A first-order Taylor expansion around the original embeddings gives a unified approximation for the entire sequence:
\begin{align}
\notag
& F(\mathbf{e}_1 + \Delta \mathbf{e}_1, \dots, \mathbf{e}_T + \Delta \mathbf{e}_T) \\
\approx & 
F(\mathbf{e}_1, \dots, \mathbf{e}_T)
+
\sum_{t=1}^{T} \nabla_{\mathbf{e}_t} F(\mathbf{e}_1, \dots, \mathbf{e}_T)^{\top} \Delta \mathbf{e}_t
\end{align}
Choosing the perturbation direction as $\Delta \mathbf{e}_t = -\mathbf{e}_t$ and rearrange the above equation, we obtain:
\begin{align}
\notag
& F(\mathbf{e}_1, \dots, \mathbf{e}_T) - F(\mathbf{0}, \dots, \mathbf{0}) \\
\approx &
\sum_{t=1}^{T} \nabla_{\mathbf{e}_t} F(\mathbf{e}_1, \dots, \mathbf{e}_T)^{\top} \mathbf{e}_t
=
\sum_{t=1}^{T} \mathbf{g}_t^\top \mathbf{e}_t
\end{align}
Each term $\mathbf{g}_t^\top \mathbf{e}_t$ measures the first-order contribution of token $a_t$ to the Judge's decision, and the sum over all tokens approximates the total change in the Judge’s output compared to its output when the input embedding is set to zero. Hence, each token's contribution can naturally be used to define the token-level reward in RL optimization, since under the first-order approximation, the sum of all token-level contributions approximates the sequence-level reward relative to a constant reference baseline. Notably, this baseline can, in principle, be chosen \citep{sundararajan2017axiomatic}. For simplicity, we set it as the zero embedding.

\subsection{Policy Optimization via Token-level GRPO}

Common RL optimization methods such as GRPO \citep{shao2024deepseekmath}, DAPO \citep{yu2025dapo}, and RLOO \citep{ahmadian2024back} rely on sequence-level rewards, where all tokens within a generated response share the same advantage signal.
This leads to coarse granularity issue \citep{sun2025ktae}.
To fully exploit the token-level rewards provided by our \method, we introduce token-level GRPO, a principled extension of GRPO that enables fine-grained optimization at the token level.

For a given input query $x$, the policy generates $G$ responses $\{o_i\}_{i=1}^{G}$, where each response
$o_i = (a_{i,1}, \dots, a_{i,T})$
is associated with token-level rewards $\{r_{i,t}\}_{t=1}^{T}$ obtained via our \method.
The return $R_{i,t}$ aggregates future token rewards from position $t$ to the end of the response.
Following standard GRPO, we compute the token-level advantage $\hat A_{i,t}$ within each group:
\begin{equation}
\hat A_{i,t}
=
\frac{
R_{i,t}
-
\mathrm{mean}\!\left(
\{\, R_{j,s} \,\}_{j=1,s=1}^{G,\;|o_j|}
\right)
}{
\mathrm{std}\!\left(
\{\, R_{j,s} \,\}_{j=1,s=1}^{G,\;|o_j|}
\right)
},
\quad
R_{i,t}
=
\sum_{k=t}^{T} r_{i,k}
\label{eq:token_adv}
\end{equation}
The optimization objective of token-level GRPO is:
\begin{multline}
\mathcal{J}(\theta)
=
\mathbb{E}_{q \sim \mathcal{D},\, \{o_i\}_{i=1}^{G} \sim \pi_{\theta_{\text{old}}}(\cdot \mid q)}
\\
\Bigg[
\frac{1}{G}
\sum_{i=1}^{G}
\frac{1}{|o_i|}
\sum_{t=1}^{|o_i|}
\min \Big(
\rho_{i,t}(\theta)\, \hat{A}_{i,t},\,
\\
\mathrm{clip}\!\left(\rho_{i,t}(\theta),\, 1-\epsilon,\, 1+\epsilon\right) \hat{A}_{i,t}
\Big)
\Bigg]
\label{eq:token_grpo}
\end{multline}
where
$\rho_{i,t}(\theta)=
\frac{
\pi_\theta(o_{i,t} \mid q, o_{i,<t})
}{
\pi_{\theta_{\text{old}}}(o_{i,t} \mid q, o_{i,<t})
}$ 
is the token-level importance ratio between the current policy and the previous policy, and $\epsilon$ is the clipping coefficient.

Alg.~\ref{alg:pseudocode} outlines the step-by-step procedure of \method. 
Notably, since each query is typically associated with $K$ rubric items $(c_k, w_k)$ that define evaluation criteria from multiple perspectives, we explicitly incorporate this practical setting into Alg.~\ref{alg:pseudocode}, although it is abstracted away in the method description for simplicity.

\begin{algorithm}[t]
\caption{Policy Optimization for Open-Ended Tasks via \method}
\label{alg:pseudocode}
\begin{algorithmic}[1]
\REQUIRE Policy $\pi_\theta$, Judge $p_{\text{judge}}$, dataset $\mathcal{D}$, rubric set $\mathcal{R}$, group size $G$, temperature $\tau$
\STATE Sample query $x \sim \mathcal{D}$
\STATE Generate a group of responses $\{o_i\}_{i=1}^{G} \sim \pi_{\theta_{\mathrm{old}}}(\cdot \mid x)$
\FOR{$i = 1$ to $G$}
    \STATE Rubric for $x$: $\mathcal{R}(x) = \{(c_k, w_k)\}_{k=1}^{K}$
    \FOR{$k = 1$ to $K$}
        \IF{$\mathbb{I}[z \sim p_{\text{judge}}(\,\cdot \mid x, o_i, c_k)] = 1$}
            \STATE Compute token-level gradients:
            \STATE $\mathbf{g}_{k,t} = \nabla_{\mathbf{e}_t}\log p_{\text{judge}}(z \mid x,o_i,c_k)$
            \STATE Convert gradients to attribution score:
            \STATE $b_{k,t} = \mathbf{g}_{k,t}^{\top}\mathbf{e}_t,\quad
                   \alpha_{k,t} = \mathrm{softmax}_t(b_{k,t}/\tau)$
        \ENDIF
    \ENDFOR
    \STATE Compute token rewards:
    \STATE $r_{i,t}
        =
        \dfrac{
            \sum_{k=1}^{K} w_k \alpha_{k,t}
        }{
            \sum_{k=1}^{K} \max(w_k,0)
        }$
\ENDFOR
\STATE Compute token-level returns $R_{i,t}$ and advantages $\hat A_{i,t}$ following Eq.~\eqref{eq:token_adv} using token rewards.
\STATE Update $\theta$ via token-level GRPO objective in Eq.~\eqref{eq:token_grpo}
\end{algorithmic}
\end{algorithm}

\section{Experiments}
\subsection{Experimental Settings}
\paragraph{Policy training.}
We train models of different families and parameter scales, including Qwen2.5-1.5B-Instruct, Qwen2.5-3B-Instruct, and Llama-3.1-8B-Instruct, using full-parameter fine-tuning.  
Detailed training procedures and hyperparameter settings are provided in Appendix~\ref{appendix:training_details}.

\paragraph{Datasets.}
For the medical consultation domain, we employ the HealthBench \citep{arora2025healthbench} and RaR-Medicine \citep{gunjal2025rubrics} datasets.
For academic question answering, we use ResearchQA \citep{yifei2025researchqa} and RaR-Science \citep{gunjal2025rubrics}.
Each dataset is split into a training set and a test set.
Each query in these datasets is associated with multiple rubric items, which define query-specific evaluation criteria for assessing the quality of LLM-generated responses.
A detailed description of these datasets is provided in Appendix \ref{appendix:datasets}.

\paragraph{Evaluation.}
To ensure fair and reliable evaluation, we adopt the OpenAI Simple-Evals suite\footnote{https://github.com/openai/simple-evals}, which computes rubric-based scores and supports pluggable grader models. 
For testing, we use Qwen3-30B-A3B-Instruct as the primary test grader to compute the average score.
Notably, the test grader is a stronger external model than the Judge used during training and is employed solely to assess the performance of the trained policy.
To demonstrate the effectiveness of training, we employ different prompt templates for training and testing. Detailed templates are in Appendix~\ref{appendix:Judge_prompt}.

\paragraph{Baselines.}
We compare our proposed method with the following baselines:
\emph{Vanilla-GRPO}, which optimizes the policy using the sequence-level reward defined in Eq.~\eqref{eq:outcome_reward}.
\emph{RuscaRL} \citep{zhou2025breaking}, which also uses sequence-level rewards but leverages rubric to guide RL exploration, representing the current leading approach for optimizing LLMs on open-ended tasks.

\subsection{Main Results}
\begin{table*}[t]
    \centering
    \caption{Main results comparison across different open-ended tasks. The best results are highlighted in \textbf{bold}.}
    \label{tab:judge_results}
    \resizebox{0.95\textwidth}{!}{
    \begin{tabular}{lcccccccc}
        \toprule
        \multirow{3}{*}{\textbf{Model}} 
        & \multicolumn{4}{c}{\textbf{Qwen3-30B-A3B-Instruct}} 
        & \multicolumn{4}{c}{\textbf{Mistral-Small-3.2-24B-Instruct}} \\
        \cmidrule(lr){2-5}\cmidrule(lr){6-9}
        & \textbf{\makecell{Health\\Bench}} & \textbf{\makecell{RaR-\\Medicine}} & \textbf{\makecell{Research\\QA}} & \textbf{\makecell{RaR-\\Science}}
        & \textbf{\makecell{Health\\Bench}} & \textbf{\makecell{RaR-\\Medicine}} & \textbf{\makecell{Research\\QA}} & \textbf{\makecell{RaR-\\Science}} \\
        \midrule
        Qwen2.5-1.5B-Instruct       & 32.2 & 27.7 & 41.2 & 36.2 & 28.1 & 29.5 & 41.5 & 48.8 \\
        \hspace*{1em}- Vanilla-GRPO  & 39.5 & 32.7 & 53.1 & 41.6 & 32.4 & 34.4 & 48.8 & 49.8  \\
        \hspace*{1em}- RuscaRL      & 40.7 & 34.3 & 53.9 & \textbf{44.1} & 32.7 & 35.8 & 49.4 & 52.0 \\
        \rowcolor{blue!6}
        \hspace*{1em}- \textbf{\method (ours)}  & \textbf{44.5} & \textbf{35.5} & \textbf{55.0} & 43.5 & \textbf{36.4} & \textbf{37.2} & \textbf{51.0} & \textbf{53.1} \\
        \midrule
        Llama-3.2-3B-Instruct       & 44.4 & 45.3 & 59.9 & 40.0 & 37.5 & 47.4 & 57.2 & 55.3 \\
        \hspace*{1em}- Vanilla-GRPO  & 46.2 & 48.0 & 63.2 & 41.1 & 38.7 & 49.7 & 58.8 & 56.5 \\
        \hspace*{1em}- RuscaRL      & 47.4 & \textbf{50.5} & 63.5 & 41.4 & 39.9 & \textbf{51.8} & \textbf{59.4} & 55.9 \\
        \rowcolor{blue!6}
        \hspace*{1em}- \textbf{\method (ours)}  & \textbf{49.4} & 49.8 & \textbf{63.6} & \textbf{43.6} & \textbf{41.7} & 51.6 & 59.2 & \textbf{58.1} \\
        \midrule
        Llama-3.1-8B-Instruct       & 45.5 & 54.9 & 63.3 & 53.6 & 39.3 & 57.0 & 60.6 & 63.3 \\
        \hspace*{1em}- Vanilla-GRPO  & 47.8 & 56.7 & 65.9 & 54.5 & 39.6 & 59.7 & 63.0 & 64.1 \\
        \hspace*{1em}- RuscaRL      & 48.6 & 61.1 & 67.0 & 56.2 & 40.6 & 60.7 & 64.3 & 65.0 \\
        \rowcolor{blue!6}
        \hspace*{1em}- \textbf{\method (ours)}  & \textbf{51.1} & \textbf{62.1} & \textbf{68.9} & \textbf{56.7} & \textbf{42.0} & \textbf{61.5} & \textbf{65.0} & \textbf{65.8} \\
        \bottomrule
    \end{tabular}
    }
\end{table*}

\begin{figure*}[t]
\centering
\includegraphics[width=0.99\linewidth]{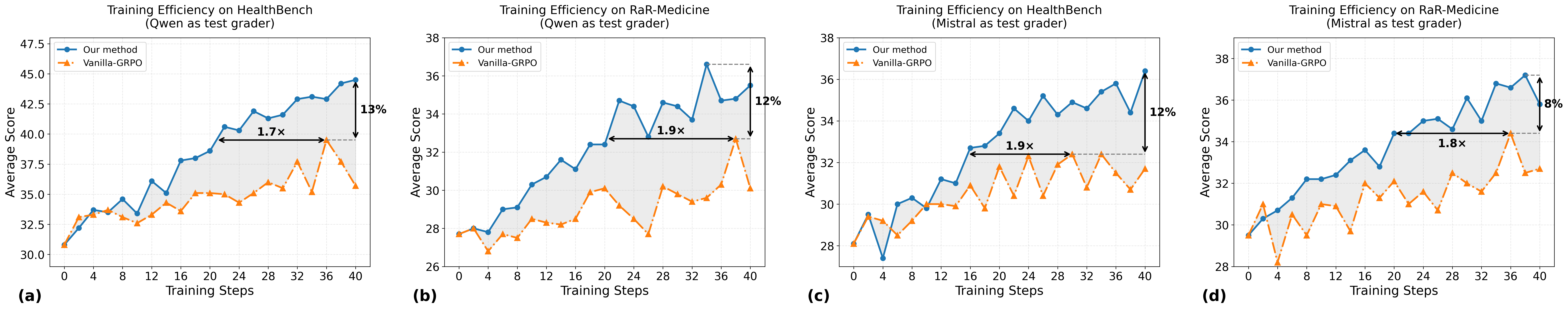}
\caption{Training dynamics on \textbf{(a)} HealthBench, \textbf{(b)} RaR-Medicine with Qwen3-30B-A3B-Instrcut as test grader, and \textbf{(c)} HealthBench, \textbf{(d)} RaR-Medicine with Mistral-Small-3.2-24B-Instruct as test grader.}
\label{fig:training_efficiency}
\end{figure*}

\paragraph{\method consistently performs well across different models and tasks.} 
Table~\ref{tab:judge_results} presents the main results across four diverse open-ended tasks, spanning different model families and sizes.
As shown, \method consistently outperforms the two sparse-reward baselines: Vanilla-GRPO and RuscaRL.
Specifically, when using Qwen2.5-1.5B-Instruct with test grader Qwen3-30B-A3B, our \method outperforms the strong baseline RuscaRL by margins of 3.8 points on HealthBench. 
Similarly, on ResearchQA,
it achieves a score of 55.0, surpassing RuscaRL’s 53.9. This indicates that dense supervision effectively compensates for the limited reasoning capacity of smaller models.
With Llama-3.1-8B-Instruct, our \method also achieves a score of 68.9 on ResearchQA, significantly outperforming Vanilla-GRPO (65.9) and RuscalRL (67.0). 
These results suggest that dense gradient-based supervision provides more informative optimization signals than sparse rewards.

\paragraph{\method is robust to different test graders.}
To rule out the possibility of overfitting to a specific evaluator, we conducted assessments using two distinct, high-capability test graders: Qwen3-30B-A3B-Instruct and Mistral-Small-3.2-24B-Instruct. The results show that \method consistently delivers strong performance across both graders. Specifically, with Qwen3-30B-A3B-Instruct as the test grader, \method achieves substantial gains across all policy models. For example, when Llama-3.2-3B-Instruct is used as the policy, \method improves HealthBench from 46.2 (Vanilla-GRPO) and 47.4 (RuscaRL) to 49.4, and boosts RaR-Science to 43.6. Similarly, with Mistral-Small-3.2-24B-Instruct, \method again outperforms the baselines: Llama-3.2-3B-Instruct trained with \method attains 41.7 on HealthBench and 58.1 on RaR-Science, demonstrating the robustness and consistency of our approach across different graders. These results underscore the adaptability and reliability of \method, highlighting its broad applicability in real-world open-ended tasks.

\subsection{Training Efficiency Analysis}
\paragraph{\method achieves higher training efficiency and stronger performance.}
Figure~\ref{fig:training_efficiency} compares the training dynamics of \method and Vanilla-GRPO on HealthBench and RaR-Medicine, evaluated with two test grader, i.e., Qwen3-30B-A3B-Instruct and Mistral-Small-3.2-24B-Instruct.
In terms of convergence speed, \method reaches the same or better performance levels as Vanilla-GRPO with 1.7×–1.9× fewer training steps under the Qwen grader and 1.8×–1.9× fewer steps under the Mistral grader.
This acceleration indicates that dense gradient-based rewards significantly accelerate optimization by providing more informative credit assignment. 
Beyond faster convergence, \method also achieves higher asymptotic performance: it yields 13\% and 12\% relative gains on HealthBench and RaR-Medicine when evaluated by Qwen, and maintains robust improvements of 12\% and 8\% under the Mistral grader. 
The consistency of these trends across datasets and test graders suggests that the advantages of \method stem from its ability to deliver fine-grained supervision, which improves both optimization efficiency and policy output quality.

\subsection{Ablation Study}
\begin{table}[t]
    \centering
    \footnotesize
    \setlength{\tabcolsep}{4pt}
    \caption{Ablation study on different gradient attribution methods.}
    \label{tab:grad_attr}
    \resizebox{\linewidth}{!}{
    \begin{tabular}{lcccc}
        \toprule
        & \makecell{Health\\Bench} & \makecell{RaR-\\Medicine} & \makecell{Research\\QA} & \makecell{RaR-\\Science}  \\
        \midrule
        Vanilla-GRPO & 32.2 & 27.7 & 41.2 & 36.2 \\
        \midrule
        L1 norm & 38.7 & 33.1 & \textbf{55.6} & 37.8 \\
        L2 norm & 38.0 & 34.5 & 54.9 & 40.0 \\
        \rowcolor{blue!6}
        \textbf{\makecell{\method (ours)}} & \textbf{44.5} & \textbf{35.5} & 55.0 & \textbf{43.5} \\
        \bottomrule
    \end{tabular}
    }
\end{table}

To assess both the effectiveness and the necessity of our attribution-based reward design, we conduct an ablation study comparing different gradient attribution strategies. The corresponding results are reported in Table~\ref{tab:grad_attr}. 
Specifically, we contrast our proposed \emph{Gradient $\times$ Embedding} formulation with magnitude-based alternatives, namely the $L_1$ and $L_2$ norms of gradients. 
Implementation details and formal definitions are deferred to Appendix~\ref{appendix:ablation_details}. 
We use Qwen2.5-1.5B-Instruct as the policy and the training-time judge, and evaluate performance with Qwen3-30B-A3B-Instruct as the test grader.

Compared to Vanilla-GRPO, which relies solely on sequence-level rewards, all gradient-assisted variants (using $L_1, L_2$ norms) consistently deliver notable improvements across benchmarks.
These results provide strong empirical evidence that dense, token-level supervision derived from gradients offers richer and more informative learning signals, enabling more effective policy optimization than sparse, sequence-level rewards.
While both $L_1$ and $L_2$ norms already outperform sparse-reward baselines, our \emph{Gradient $\times$ Embedding} approach consistently achieves the best performance—for example, improving HealthBench from 38.0 to 44.5 and RaR-Science from 40.0 to 43.5.
These results indicate that our method provides more accurate token-level rewards, leading to more effective policy optimization.

As shown in Section~\ref{sec:theoretical_analysis}, \emph{Gradient $\times$ Embedding} naturally arises from the first-order Taylor expansion of the Judge’s decision function, directly approximating each token’s contribution to the Judge's verdict.
In contrast, norm-based metrics capture only gradient magnitude and discard directional information. 
By preserving gradient directionality, our formulation explicitly ensures that token-level rewards remain mathematically consistent with the global optimization objective, which is essential for dense supervision.

\subsection{Extended Analysis}
In this section, we conduct a more in-depth analysis of our method by investigating the following research questions.
\textbf{RQ1}: How does the performance of our self-judging mechanism compare with that of a more capable external Judge model?
\textbf{RQ2}: Can our \method be extended to other RL optimization methods?
\textbf{RQ3}: How well does a policy trained with \method on one dataset generalize to other datasets?
\textbf{RQ4}: How does our gradient-based reward attribution method perform compared with latest developed process reward models?

\subsubsection{Analysis of the Self-Judging Mechanism}
\begin{table}[t]
    \centering
    \footnotesize
    \setlength{\tabcolsep}{4pt}
    \caption{Performance of different Judge used for training when the policy uses Qwen2.5-1.5B-Instruct.}
    \label{tab:train_judge}
    \resizebox{\linewidth}{!}{
    \begin{tabular}{lcccc}
        \toprule
        Judge for training & \makecell{Health\\Bench} & \makecell{RaR-\\Medicine} & \makecell{Research\\QA} & \makecell{RaR-\\Science}  \\
        \midrule
        \makecell{Qwen2.5-1.5B-Instruct\\ Self-judging}  & 44.5 & 35.5 & 55.0 & 43.3 \\
        \midrule
        Qwen2.5-7B-Instruct                & 45.3 & 36.6 & 55.6 & 43.1 \\
        Llama3.1-8B-Instruct               & 43.7 & 34.5 & 56.8 & 43.6 \\
        Qwen3-30B                          & 45.9 & 36.3 & 55.4 & 43.5 \\
        \bottomrule
    \end{tabular}
    }
\end{table}
\paragraph{Self-judging achieves competitive performance (RQ1).}
Table~\ref{tab:train_judge} compares the performance of policies trained with different Judge models while fixing the policy to Qwen2.5-1.5B-Instruct. 
Overall, using the policy itself as the judge, namely the self-judging mechanism, achieves performance that is highly competitive with, and in some cases comparable to, training with substantially larger and more capable judge models.
Across all four benchmarks, the self-judging setup consistently delivers strong results, with only marginal gaps relative to larger judges, including Qwen2.5-7B, Llama3.1-8B, and Qwen3-30B. 
Notably, on HealthBench and RaR-Medicine, self-judging performs on par with or only slightly below the best-performing external judges, while exhibiting comparable performance on ResearchQA and RaR-Science. 
These results indicate that the proposed learning signal does not critically rely on a stronger external Judge. 
Instead, \textbf{the policy model itself can provide sufficiently informative evaluative feedback for its own optimization}.

\subsubsection{\method with RLOO}
\begin{table}[t]
    \centering
    \footnotesize
    \setlength{\tabcolsep}{4pt}
    \caption{Performance of our \method against baselines under RLOO optimization.}
    \label{tab:rloo}
    \resizebox{\linewidth}{!}{
    \begin{tabular}{lcccc}
        \toprule
        & \makecell{Health\\Bench} & \makecell{RaR-\\Medicine} & \makecell{Research\\QA} & \makecell{RaR-\\Science}  \\
        \midrule
        Vanilla-RLOO  & 39.3 & 33.5 & 52.2 & 41.0 \\
        RuscaRL      & 40.5 & 32.5 & 49.3 & 42.3 \\
        \rowcolor{blue!6}
        \textbf{\method (ours)}  & \textbf{42.9} & \textbf{33.9} & \textbf{57.2} & \textbf{42.8} \\
        \bottomrule
    \end{tabular}
    }
\end{table}

\begin{table*}[t]
    \centering
    \footnotesize
    \setlength{\tabcolsep}{4pt}
    \caption{Pass@1 accuracy evaluated on six mathematical reasoning benchmarks.}
    \label{tab:dense_model_comparison}
    \resizebox{0.7\linewidth}{!}{
    \begin{tabular}{ccccccc}
        \toprule
        \textbf{Method} & \textbf{\makecell{MATH\\500}} & \textbf{\makecell{Minerva\\Math}} & \textbf{\makecell{Olympiad\\Bench}} & \textbf{\makecell{AIME25}} & \textbf{\makecell{AIME24}} & \textbf{\makecell{AMC23}}  \\
        \midrule
        PURE \citep{cheng2025stop}      & 76.0 & 30.8 & 36.7 & 13.3 & 26.6 & \textbf{70.0} \\
        PRM \citep{wang-etal-2024-math} & 71.6 & 36.3 & 32.5 & 13.3 & 10.0 & 57.5 \\
        PQM \citep{li2025process}       & 72.0 & 34.1 & 34.3 & 13.3 & 13.3 & 52.5 \\
        \rowcolor{blue!6}
        \textbf{\method (ours)}     & \textbf{77.6} & \textbf{36.7} & \textbf{38.3} & \textbf{16.6} & \textbf{26.6} & 65.0 \\
        \bottomrule
    \end{tabular}
    }
\end{table*}

\begin{table}[t]
\centering
\footnotesize
\caption{Cross-dataset generalization performance.}
\label{tab:cross_generalization}
    \resizebox{\linewidth}{!}{
    \begin{tabular}{lcccc}
    \toprule
    Training Set & Test Set & Vanilla-GRPO & RuscaRL & \textbf{Ours} \\
    \midrule
    \multirow{1}{*}{RaR-Medicine} & HealthBench & 37.6 & 37.3 & \textbf{42.1} \\
    \midrule
    \multirow{2}{*}{RaR-Science} & ResearchQA & 45.1 & 46.2 & \textbf{48.6} \\
    & GPQA-Diamond & 24.7 & 25.6 & \textbf{26.1} \\
    \bottomrule
    \end{tabular}
    }
\end{table}

\paragraph{\method remains effective under RLOO optimization (RQ2).} 
Table~\ref{tab:rloo} reports the performance of \method when integrated with the token-level RLOO algorithm, with details provided in Appendix~\ref{appendix:rloo}. 
Sparse-reward baselines adopt the original RLOO algorithm.
Despite replacing the underlying RL algorithm, \method consistently outperforms both Vanilla-RLOO and RuscaRL across all four benchmarks, demonstrating that its advantages are not tied to a specific optimization scheme.
In particular, \method achieves a substantial improvement of +5.0 points over Vanilla-RLOO and +7.9 points over RuscaRL on ResearchQA. 
Similar gains are observed on HealthBench and RaR-Science, indicating robust performance across diverse tasks.
These results indicate that the effectiveness of \method stems from the reward formulation itself and remains compatible with different policy optimization schemes.

\subsubsection{Cross-Dataset Generalization}
\paragraph{\method-trained policies exhibit stronger cross-dataset generalization on both open-ended and verifiable tasks (RQ3).}
As shown in Table \ref{tab:cross_generalization}, when trained on RaR-Medicine and evaluated on HealthBench, our \method significantly outperforms both Vanilla-GRPO and RuscaRL by 4.5 and 4.8 points. 
In the science domain, we train on RaR-Science and observe that \method generalizes better to both the open-ended benchmark ResearchQA and the verifiable benchmark GPQA-Diamond \citep{rein2024gpqa}.
GPQA-Diamond demands multi-step scientific reasoning, requires precise factual grounding, and provides verifiable ground truth answers.
Although trained only on the open-ended dataset, \method achieves a score of 26.1 on GPQA-Diamond compared to 24.7 and 25.6 from Vanilla-GRPO and RuscaRL.
This indicates that dense gradient-based reward improves the structural quality and factual fidelity of policy outputs, enabling better generalization beyond the training distribution and task format.

\subsubsection{Compared with PRMs on Mathematical Reasoning Tasks}
Existing approaches in mathematical reasoning typically rely on PRMs to supply dense supervision. 
To compare against these approaches under a unified setting, we extend \method to verifiable domains.
Detailed implementation details are provided in the Appendix~\ref{appendix:reward_model}.
We compare our \method with three representative dense reward modeling methods: PRM \citep{wang2024math}, PQM \citep{li2025process}, PURE \citep{cheng2025stop}
The performance is evaluated on six widely adopted benchmarks: MATH500 \citep{hendrycks2021measuring}, Minerva Math \citep{lewkowycz2022solving}, OlympiadBench \citep{he2024olympiadbench}, AIME24 \citep{maa2024aime}, AIME25 \citep{maa2025aime}, and AMC23 \citep{maa2023aime}. 
We report the Pass@1 accuracy under the zero-shot setting.

\paragraph{\method achieves better performance than PRMs on mathematical reasoning tasks.(RQ4)}
Table~\ref{tab:dense_model_comparison} presents the comparative results. 
Overall, \method demonstrates superior performance, achieving the highest accuracy on 5 out of 6 benchmarks.
On MATH500, \method attains 77.6\% accuracy, surpassing PURE (76.0\%) and substantially outperforming PQM (72.0\%). 
On more challenging competition-level benchmarks such as AIME24, \method matches the strong performance of PURE (26.6\%) while significantly outperforming the standard PRM baseline, suggesting that gradient-based dense rewards can provide supervision quality comparable to explicitly trained PRMs even in constrained verifiable settings.
These results demonstrate that extending \method to verifiable domains preserves the core advantages observed in open-ended tasks. 
By providing dense and informative supervision, \method generalizes effectively across benchmarks with varying reasoning complexity. 
This unified behavior positions \method as a more scalable and flexible alternative to PRM-based approaches, capable of supporting both open-ended and verifiable reasoning tasks within a single reward framework.

\section{Conclusion}
In this work, we proposed \method, which extracts dense, token-level rewards directly from the Judge’s internal gradient signals via a single backward pass. This approach addresses key limitations of prior methods, including sparse rewards and the black-box treatment of the Judge. Experimental results across diverse open-ended tasks and different policy models demonstrate that \method consistently outperforms strong baselines, achieves competitive performance through self-judging, and offers superior training efficiency. Looking forward, our method can be extended to long-horizon agent tasks involving multiple decision steps and sustained reasoning, where gradient-based rewards have the potential to provide high-quality process supervision and enable more stable policy optimization.

\section*{Impact Statement}
This paper presents work aimed at advancing the field of machine learning by improving the training efficiency and reasoning quality of Large Language Models in open-ended domains. By introducing a framework for dense, token-level credit assignment, our research offers a path toward developing highly capable AI systems in fields where ground-truth verification is difficult, such as specialized research and creative problem-solving. This approach significantly reduces the need for expensive, human-intensive labeling and the reliance on massive external reward models, potentially lowering the barriers to entry for developing sophisticated AI agents.

However, the use of self-evaluative signals in sensitive areas like medical or legal consultation necessitates careful implementation. While the proposed method enhances reasoning consistency, the societal impact depends heavily on the accuracy of the underlying models used as judges. We encourage practitioners to combine our framework with robust safety guardrails and multi-faceted evaluation protocols to ensure that generated content remains factually accurate and ethically sound. The goal is to support the creation of reliable AI assistants that can provide nuanced and scientifically grounded support to human experts.


\bibliography{grad_reward}

@article{guo2025deepseek,
  title={DeepSeek-R1 incentivizes reasoning in LLMs through reinforcement learning},
  author={Guo, Daya and Yang, Dejian and Zhang, Haowei and Song, Junxiao and Wang, Peiyi and Zhu, Qihao and Xu, Runxin and Zhang, Ruoyu and Ma, Shirong and Bi, Xiao and others},
  journal={Nature},
  volume={645},
  number={8081},
  pages={633--638},
  year={2025},
  publisher={Nature Publishing Group UK London}
}

@article{gunjal2025rubrics,
  title={Rubrics as rewards: Reinforcement learning beyond verifiable domains},
  author={Gunjal, Anisha and Wang, Anthony and Lau, Elaine and Nath, Vaskar and He, Yunzhong and Liu, Bing and Hendryx, Sean},
  journal={arXiv preprint arXiv:2507.17746},
  year={2025}
}

@article{zhou2025breaking,
  title={Breaking the exploration bottleneck: Rubric-scaffolded reinforcement learning for general llm reasoning},
  author={Zhou, Yang and Li, Sunzhu and Liu, Shunyu and Fang, Wenkai and Zhang, Kongcheng and Zhao, Jiale and Yang, Jingwen and Zhou, Yihe and Lv, Jianwei and Zheng, Tongya and others},
  journal={arXiv preprint arXiv:2508.16949},
  year={2025}
}

@article{arora2025healthbench,
  title={Healthbench: Evaluating large language models towards improved human health},
  author={Arora, Rahul K and Wei, Jason and Hicks, Rebecca Soskin and Bowman, Preston and Qui{\~n}onero-Candela, Joaquin and Tsimpourlas, Foivos and Sharman, Michael and Shah, Meghan and Vallone, Andrea and Beutel, Alex and others},
  journal={arXiv preprint arXiv:2505.08775},
  year={2025}
}

@article{jaech2024openai,
  title={Openai o1 system card},
  author={Jaech, Aaron and Kalai, Adam and Lerer, Adam and Richardson, Adam and El-Kishky, Ahmed and Low, Aiden and Helyar, Alec and Madry, Aleksander and Beutel, Alex and Carney, Alex and others},
  journal={arXiv preprint arXiv:2412.16720},
  year={2024}
}

@article{shao2024deepseekmath,
  title={Deepseekmath: Pushing the limits of mathematical reasoning in open language models},
  author={Shao, Zhihong and Wang, Peiyi and Zhu, Qihao and Xu, Runxin and Song, Junxiao and Bi, Xiao and Zhang, Haowei and Zhang, Mingchuan and Li, YK and Wu, Yang and others},
  journal={arXiv preprint arXiv:2402.03300},
  year={2024}
}

@inproceedings{
yu2025dapo,
title={{DAPO}: An Open-Source {LLM} Reinforcement Learning System at Scale},
author={Qiying Yu and Zheng Zhang and Ruofei Zhu and Yufeng Yuan and Xiaochen Zuo and YuYue and Weinan Dai and Tiantian Fan and Gaohong Liu and Juncai Liu and LingJun Liu and Xin Liu and Haibin Lin and Zhiqi Lin and Bole Ma and Guangming Sheng and Yuxuan Tong and Chi Zhang and Mofan Zhang and Ru Zhang and Wang Zhang and Hang Zhu and Jinhua Zhu and Jiaze Chen and Jiangjie Chen and Chengyi Wang and Hongli Yu and Yuxuan Song and Xiangpeng Wei and Hao Zhou and Jingjing Liu and Wei-Ying Ma and Ya-Qin Zhang and Lin Yan and Yonghui Wu and Mingxuan Wang},
booktitle={The Thirty-ninth Annual Conference on Neural Information Processing Systems},
year={2025},
url={https://openreview.net/forum?id=2a36EMSSTp}
}

@article{bi2025reward,
  title={Reward and Guidance through Rubrics: Promoting Exploration to Improve Multi-Domain Reasoning},
  author={Bi, Baolong and Liu, Shenghua and Wang, Yiwei and Tong, Siqian and Mei, Lingrui and Ge, Yuyao and Xu, Yilong and Guo, Jiafeng and Cheng, Xueqi},
  journal={arXiv preprint arXiv:2511.12344},
  year={2025}
}

@article{sheng2024hybridflow,
  title   = {HybridFlow: A Flexible and Efficient RLHF Framework},
  author  = {Guangming Sheng and Chi Zhang and Zilingfeng Ye and Xibin Wu and Wang Zhang and Ru Zhang and Yanghua Peng and Haibin Lin and Chuan Wu},
  year    = {2024},
  journal = {arXiv preprint arXiv: 2409.19256}
}

@article{shao2025dr,
  title={Dr tulu: Reinforcement learning with evolving rubrics for deep research},
  author={Shao, Rulin and Asai, Akari and Shen, Shannon Zejiang and Ivison, Hamish and Kishore, Varsha and Zhuo, Jingming and Zhao, Xinran and Park, Molly and Finlayson, Samuel G and Sontag, David and others},
  journal={arXiv preprint arXiv:2511.19399},
  year={2025}
}

@article{yifei2025researchqa,
  title={Researchqa: Evaluating scholarly question answering at scale across 75 fields with survey-mined questions and rubrics},
  author={Yifei, Li S and Chang, Allen and Malaviya, Chaitanya and Yatskar, Mark},
  journal={arXiv preprint arXiv:2509.00496},
  year={2025}
}

@article{wang2025infimed,
  title={InfiMed-ORBIT: Aligning LLMs on Open-Ended Complex Tasks via Rubric-Based Incremental Training},
  author={Wang, Pengkai and Liu, Pengwei and Sang, Zhijie and Xie, Congkai and Yang, Hongxia and others},
  journal={arXiv preprint arXiv:2510.15859},
  year={2025}
}

@inproceedings{
sun2025ktae,
title={{KTAE}: A Model-Free Algorithm to Key-Tokens Advantage Estimation in Mathematical Reasoning},
author={Wei Sun and Wen Yang and Pu Jian and Qianlong Du and Fuwei Cui and Shuo Ren and Jiajun Zhang},
booktitle={The Thirty-ninth Annual Conference on Neural Information Processing Systems},
year={2025},
url={https://openreview.net/forum?id=yqQVRNdmKJ}
}

@article{cui2025process,
  title={Process reinforcement through implicit rewards},
  author={Cui, Ganqu and Yuan, Lifan and Wang, Zefan and Wang, Hanbin and Zhang, Yuchen and Chen, Jiacheng and Li, Wendi and He, Bingxiang and Fan, Yuchen and Yu, Tianyu and others},
  journal={arXiv preprint arXiv:2502.01456},
  year={2025}
}

@article{zhang2025linking,
  title={Linking Process to Outcome: Conditional Reward Modeling for LLM Reasoning},
  author={Zhang, Zheng and Shan, Ziwei and Song, Kaitao and Li, Yexin and Ren, Kan},
  journal={arXiv preprint arXiv:2509.26578},
  year={2025}
}

@inproceedings{
cheng2025stop,
title={Stop Summation: Min-Form Credit Assignment Is All Process Reward Model Needs for Reasoning},
author={Jie Cheng and Gang Xiong and Ruixi Qiao and Lijun Li and Chao Guo and Junle Wang and Yisheng Lv and Fei-Yue Wang},
booktitle={The Thirty-ninth Annual Conference on Neural Information Processing Systems},
year={2025},
url={https://openreview.net/forum?id=3Sxby0hH1q}
}

@inproceedings{wang-etal-2024-math,
    title = "Math-Shepherd: Verify and Reinforce {LLM}s Step-by-step without Human Annotations",
    author = "Wang, Peiyi  and
      Li, Lei  and
      Shao, Zhihong  and
      Xu, Runxin  and
      Dai, Damai  and
      Li, Yifei  and
      Chen, Deli  and
      Wu, Yu  and
      Sui, Zhifang",
    editor = "Ku, Lun-Wei  and
      Martins, Andre  and
      Srikumar, Vivek",
    booktitle = "Proceedings of the 62nd Annual Meeting of the Association for Computational Linguistics (Volume 1: Long Papers)",
    month = aug,
    year = "2024",
    address = "Bangkok, Thailand",
    publisher = "Association for Computational Linguistics",
    url = "https://aclanthology.org/2024.acl-long.510/",
    doi = "10.18653/v1/2024.acl-long.510",
    pages = "9426--9439"
}

@inproceedings{
zeng2025versaprm,
title={Versa{PRM}: Multi-Domain Process Reward Model via Synthetic Reasoning Data},
author={Thomas Zeng and Shuibai Zhang and Shutong Wu and Christian Classen and Daewon Chae and Ethan Ewer and Minjae Lee and Heeju Kim and Wonjun Kang and Jackson Kunde and Ying Fan and Jungtaek Kim and Hyung Il Koo and Kannan Ramchandran and Dimitris Papailiopoulos and Kangwook Lee},
booktitle={Forty-second International Conference on Machine Learning},
year={2025},
url={https://openreview.net/forum?id=l19DmXbwPK}
}

@article{chen2025judgelrm,
  title={Judgelrm: Large reasoning models as a judge},
  author={Chen, Nuo and Hu, Zhiyuan and Zou, Qingyun and Wu, Jiaying and Wang, Qian and Hooi, Bryan and He, Bingsheng},
  journal={arXiv preprint arXiv:2504.00050},
  year={2025}
}

@inproceedings{lee2025checkeval,
  title={Checkeval: A reliable llm-as-a-judge framework for evaluating text generation using checklists},
  author={Lee, Yukyung and Kim, Joonghoon and Kim, Jaehee and Cho, Hyowon and Kang, Jaewook and Kang, Pilsung and Kim, Najoung},
  booktitle={Proceedings of the 2025 Conference on Empirical Methods in Natural Language Processing},
  pages={15782--15809},
  year={2025}
}

@inproceedings{wu2025meta,
  title={Meta-rewarding language models: Self-improving alignment with llm-as-a-meta-judge},
  author={Wu, Tianhao and Yuan, Weizhe and Golovneva, Olga and Xu, Jing and Tian, Yuandong and Jiao, Jiantao and Weston, Jason E and Sukhbaatar, Sainbayar},
  booktitle={Proceedings of the 2025 Conference on Empirical Methods in Natural Language Processing},
  pages={11548--11565},
  year={2025}
}

@article{xu2025direct,
  title={Direct reasoning optimization: Llms can reward and refine their own reasoning for open-ended tasks},
  author={Xu, Yifei and Chakraborty, Tusher and Sharma, Srinagesh and Nunes, Leonardo and K{\i}c{\i}man, Emre and Lu, Songwu and Chandra, Ranveer},
  journal={arXiv preprint arXiv:2506.13351},
  year={2025}
}

@article{ye2025self,
  title={Self-Rewarding Rubric-Based Reinforcement Learning for Open-Ended Reasoning},
  author={Ye, Zhiling and Yue, Yun and Wang, Haowen and Han, Xudong and Jiang, Jiadi and Wei, Cheng and Fan, Lei and Liang, Jiaxin and Zhang, Shuowen and Li, Ji and others},
  journal={arXiv preprint arXiv:2509.25534},
  year={2025}
}

@article{huang2025reinforcement,
  title={Reinforcement learning with rubric anchors},
  author={Huang, Zenan and Zhuang, Yihong and Lu, Guoshan and Qin, Zeyu and Xu, Haokai and Zhao, Tianyu and Peng, Ru and Hu, Jiaqi and Shen, Zhanming and Hu, Xiaomeng and others},
  journal={arXiv preprint arXiv:2508.12790},
  year={2025}
}

@inproceedings{
lightman2024lets,
title={Let's Verify Step by Step},
author={Hunter Lightman and Vineet Kosaraju and Yuri Burda and Harrison Edwards and Bowen Baker and Teddy Lee and Jan Leike and John Schulman and Ilya Sutskever and Karl Cobbe},
booktitle={The Twelfth International Conference on Learning Representations},
year={2024},
url={https://openreview.net/forum?id=v8L0pN6EOi}
}

@inproceedings{wang2024math,
  title={Math-shepherd: Verify and reinforce llms step-by-step without human annotations},
  author={Wang, Peiyi and Li, Lei and Shao, Zhihong and Xu, Runxin and Dai, Damai and Li, Yifei and Chen, Deli and Wu, Yu and Sui, Zhifang},
  booktitle={Proceedings of the 62nd Annual Meeting of the Association for Computational Linguistics (Volume 1: Long Papers)},
  pages={9426--9439},
  year={2024}
}

@inproceedings{
li2025process,
title={Process Reward Model with Q-value Rankings},
author={Wendi Li and Yixuan Li},
booktitle={The Thirteenth International Conference on Learning Representations},
year={2025},
url={https://openreview.net/forum?id=wQEdh2cgEk}
}

@inproceedings{
hu2025openreasonerzero,
title={Open-Reasoner-Zero: An Open Source Approach to Scaling Up Reinforcement Learning on the Base Model},
author={Jingcheng Hu and Yinmin Zhang and Qi Han and Daxin Jiang and Xiangyu Zhang and Heung-Yeung Shum},
booktitle={The Thirty-ninth Annual Conference on Neural Information Processing Systems},
year={2025},
url={https://openreview.net/forum?id=NFM8F5cV0V}
}

@misc{maa2023aime,
  author       = {{MAA}},
  title        = {American Invitational Mathematics Examination (AIME)},
  year         = {2023},
  howpublished = {\url{https://maa.org/}},
}

@misc{maa2024aime,
  author       = {{MAA}},
  title        = {American Invitational Mathematics Examination (AIME)},
  year         = {2024},
  howpublished = {\url{https://maa.org/}},
}

@misc{maa2025aime,
  author       = {{MAA}},
  title        = {American Invitational Mathematics Examination (AIME)},
  year         = {2025},
  howpublished = {\url{https://maa.org/}},
}

@inproceedings{hendrycks2021measuring,
    title={Measuring Mathematical Problem Solving With the {MATH} Dataset},
    author={Dan Hendrycks and Collin Burns and Saurav Kadavath and Akul Arora and Steven Basart and Eric Tang and Dawn Song and Jacob Steinhardt},
    booktitle={Thirty-fifth Conference on Neural Information Processing Systems Datasets and Benchmarks Track (Round 2)},
    year={2021},
}

@article{lewkowycz2022solving,
  title={Solving quantitative reasoning problems with language models},
  author={Lewkowycz, Aitor and Andreassen, Anders and Dohan, David and Dyer, Ethan and Michalewski, Henryk and Ramasesh, Vinay and Slone, Ambrose and Anil, Cem and Schlag, Imanol and Gutman-Solo, Theo and others},
  journal={Advances in Neural Information Processing Systems},
  volume={35},
  pages={3843--3857},
  year={2022}
}

@article{zhang2025lessons,
  title={The lessons of developing process reward models in mathematical reasoning},
  author={Zhang, Zhenru and Zheng, Chujie and Wu, Yangzhen and Zhang, Beichen and Lin, Runji and Yu, Bowen and Liu, Dayiheng and Zhou, Jingren and Lin, Junyang},
  journal={arXiv preprint arXiv:2501.07301},
  year={2025}
}

@inproceedings{he2024olympiadbench,
    title = "{O}lympiad{B}ench: A Challenging Benchmark for Promoting {AGI} with Olympiad-Level Bilingual Multimodal Scientific Problems",
    author = "He, Chaoqun  and
      Luo, Renjie  and
      Bai, Yuzhuo  and
      Hu, Shengding  and
      Thai, Zhen  and
      Shen, Junhao  and
      Hu, Jinyi  and
      Han, Xu  and
      Huang, Yujie  and
      Zhang, Yuxiang  and
      Liu, Jie  and
      Qi, Lei  and
      Liu, Zhiyuan  and
      Sun, Maosong",
    editor = "Ku, Lun-Wei  and
      Martins, Andre  and
      Srikumar, Vivek",
    booktitle = "Proceedings of the 62nd Annual Meeting of the Association for Computational Linguistics (Volume 1: Long Papers)",
    month = aug,
    year = "2024",
    address = "Bangkok, Thailand",
    publisher = "Association for Computational Linguistics",
    url = "https://aclanthology.org/2024.acl-long.211/",
    doi = "10.18653/v1/2024.acl-long.211",
    pages = "3828--3850",
}

@inproceedings{rein2024gpqa,
  title={Gpqa: A graduate-level google-proof q\&a benchmark},
  author={Rein, David and Hou, Betty Li and Stickland, Asa Cooper and Petty, Jackson and Pang, Richard Yuanzhe and Dirani, Julien and Michael, Julian and Bowman, Samuel R},
  booktitle={First Conference on Language Modeling},
  year={2024}
}

@inproceedings{guan2025rstar,
    title={rStar-Math: Small {LLM}s Can Master Math Reasoning with Self-Evolved Deep Thinking},
    author={Xinyu Guan and Li Lyna Zhang and Yifei Liu and Ning Shang and Youran Sun and Yi Zhu and Fan Yang and Mao Yang},
    booktitle={Forty-second International Conference on Machine Learning},
    year={2025}
}

@article{cobbe2021training,
  title={Training verifiers to solve math word problems},
  author={Cobbe, Karl and Kosaraju, Vineet and Bavarian, Mohammad and Chen, Mark and Jun, Heewoo and Kaiser, Lukasz and Plappert, Matthias and Tworek, Jerry and Hilton, Jacob and Nakano, Reiichiro and others},
  journal={arXiv preprint arXiv:2110.14168},
  year={2021}
}

@inproceedings{ahmadian2024back,
    title = "Back to Basics: Revisiting {REINFORCE}-Style Optimization for Learning from Human Feedback in {LLM}s",
    author = {Ahmadian, Arash  and
      Cremer, Chris  and
      Gall{\'e}, Matthias  and
      Fadaee, Marzieh  and
      Kreutzer, Julia  and
      Pietquin, Olivier  and
      {\"U}st{\"u}n, Ahmet  and
      Hooker, Sara},
    editor = "Ku, Lun-Wei  and
      Martins, Andre  and
      Srikumar, Vivek",
    booktitle = "Proceedings of the 62nd Annual Meeting of the Association for Computational Linguistics (Volume 1: Long Papers)",
    month = aug,
    year = "2024",
    address = "Bangkok, Thailand",
    publisher = "Association for Computational Linguistics",
    url = "https://aclanthology.org/2024.acl-long.662/",
    doi = "10.18653/v1/2024.acl-long.662",
    pages = "12248--12267",
}

@inproceedings{sundararajan2017axiomatic,
  title={Axiomatic attribution for deep networks},
  author={Sundararajan, Mukund and Taly, Ankur and Yan, Qiqi},
  booktitle={International conference on machine learning},
  pages={3319--3328},
  year={2017},
  organization={PMLR}
}

@inproceedings{yu2024ovm,
  title={Ovm, outcome-supervised value models for planning in mathematical reasoning},
  author={Yu, Fei and Gao, Anningzhe and Wang, Benyou},
  booktitle={Findings of the Association for Computational Linguistics: NAACL 2024},
  pages={858--875},
  year={2024}
}

@inproceedings{zheng2025processbench,
  title={Processbench: Identifying process errors in mathematical reasoning},
  author={Zheng, Chujie and Zhang, Zhenru and Zhang, Beichen and Lin, Runji and Lu, Keming and Yu, Bowen and Liu, Dayiheng and Zhou, Jingren and Lin, Junyang},
  booktitle={Proceedings of the 63rd Annual Meeting of the Association for Computational Linguistics (Volume 1: Long Papers)},
  pages={1009--1024},
  year={2025}
}

@inproceedings{
song2025mind,
title={Mind the Gap: Examining the Self-Improvement Capabilities of Large Language Models},
author={Yuda Song and Hanlin Zhang and Carson Eisenach and Sham M. Kakade and Dean Foster and Udaya Ghai},
booktitle={The Thirteenth International Conference on Learning Representations},
year={2025},
url={https://openreview.net/forum?id=mtJSMcF3ek}
}
\bibliographystyle{icml2026}

\newpage
\appendix
\onecolumn

\section{Implementation Details}
\label{appendix:implementation_details}

\subsection{Policy Training}
\label{appendix:training_details}
We train with a fixed learning rate of \(1\times10^{-6}\), a prompt batch size of 32, and 8 sampled responses per prompt, with the clipping ratio set to \(\epsilon = 0.2\).
Sampling is performed via vLLM with temperature \(0.7\), top-p \(0.8\), top-k \(20\), and a maximum response length of 4096 tokens.
At test time we set the temperature to 0 and top-p to 1.
All baselines and our method are implemented using the veRL~\citep{sheng2024hybridflow} framework, and all experiments are conducted on 8 NVIDIA H20 GPUs.

\subsection{Details of Gradient Attribution Baselines}
\label{appendix:ablation_details}

In the ablation study regarding gradient attribution methods, we employ the exact same pipeline as our proposed framework, with the only difference being the calculation of the importance score $a_t$. We substitute the \emph{Gradient $\times$ Embedding} calculation in \eqref{eq:gradient_attribution} with the following two formulations respectively:

\paragraph{$L_1$ norm baseline.}
We compute the importance score as the $L_1$ norm of the gradient vector $\mathbf{g}_t$:
\begin{equation}
b_{k,t} = \| \mathbf{g}_{k,t} \|_1 = \sum_{i} | g_{k,t,i} |
\end{equation}

\paragraph{$L_2$ norm baseline.}
We compute the importance score as the $L_2$ norm (Euclidean norm) of the gradient vector $\mathbf{g}_t$:
\begin{equation}
b_{k,t} = \| \mathbf{g}_{k,t} \|_2 = \sqrt{ \sum_{i} g_{k,t,i}^2 }
\end{equation}

\subsection{RLOO Optimization}
\label{appendix:rloo}
RLOO (Leave-One-Out) is an advantage estimation method based on multiple samples for optimizing policy gradients in reinforcement learning. To provide finer-grained and lower-variance advantage estimates, we use token level RLOO to construct a baseline estimate using multiple output responses excluding the current sample. In our setup, given a query, the model generates \( G \) responses, each containing at most \( M \) tokens. Let \(r_{i,j}\) denotes the reward of \( j \)-th token in the \( i \)-th response, we compute the corresponding advantage \(\hat A_{i,t}\) by subtracting the cumulative rewards by an average baseline from parallel samples: 
\begin{equation}
\label{eq:rloo_Advantage}
\hat A_{i,t}
=
\sum_{k=t}^{M} r_{i,k}
-
\frac{1}{(G-1)M}
\sum_{\substack{j=1 \\ j \neq i}}^{G}
\sum_{s=1}^{M}
\sum_{k=s}^{M}
r_{j,k}
\end{equation}

\subsection{Dataset Detail}
\label{appendix:datasets}
HealthBench \citep{arora2025healthbench} assesses LLM performance in healthcare and contains 5,000 samples, from which we select 500 instances as the test set and use the remaining samples for training. 
RaR-Medicine \citep{gunjal2025rubrics} evaluates a model’s medical question answering ability. After removing samples with duplicate questions, we partition the dataset into a training set with 17,011 samples and a test set with 500 samples.
ResearchQA \citep{yifei2025researchqa} is a large-scale benchmark for long-form scholarly question answering spanning 75 academic fields, using queries and rubrics mined from survey articles. We split the dataset into a training set with 16,961 instances and a test set with 500 instances. 
RaR-Science \citep{gunjal2025rubrics} evaluates model performance in the science domain. After removing duplicate-question samples as well, we divide the dataset into a training set with 16,365 samples and a test set with 500 samples.

\subsection{Extending \method to Verifiable Domains}
\label{appendix:reward_model}
\textbf{PRMs Baseline Training.}
The reward models are initialized from Qwen2.5-Math-7B and trained on the Math-Shepherd \citep{wang2024math} dataset, which integrates questions from GSM8K \citep{cobbe2021training} and MATH \citep{hendrycks2021measuring} and provides responses annotated with step-level process labels. 
To separate reasoning steps, we follow prior work \citep{yu2024ovm, zheng2025processbench, zhang2025lessons}, using newline characters to indicate boundaries between individual steps.
Following prior work \citep{li2025process, guan2025rstar}, we augment the pre-trained model with an additional value head to directly output scalar rewards. 
We used the ZeRO-2 optimization stage of DeepSpeed with bfloat16 precision to train the model, and employed AdamW optimizer with a learning rate of 5e-6 and a batch size of 32.


\paragraph{Combining \method with an Outcome Reward Model.}
When applying \method to mathematical domains, we employ an outcome reward model (ORM) and perform gradient attribution on it.
We show the full algorithm when combining \method with discriminative ORMs. 
Unlike the generative judge setting where the target is a decision token $z$, an ORM $V$ maps the input $x$ and response $o$ directly to a scalar score $V(x, o)$, which reflects the ORM's assessment of the response's correctness. To extract dense rewards in this setting, we modify the gradient attribution objective to this scalar output. The gradient calculation in Eq.~\eqref{eq:token_gradient} becomes:
\begin{equation}
    \mathbf{g_t} = \nabla_{\mathbf{e}_t} V(x, o)
\end{equation}
The attribution score $a_t$ and normalized reward $r_t$ are computed following the same procedure described in Section~\ref{sec:via_gradient_attribution}. Crucially, to fully leverage the global oversight of the ORM, we incorporate the final outcome score into the reward of the last token: \begin{equation}
    r_t = \begin{cases}
        r_t & \text{if } t < T \\
        r_t + V(x, o) & \text{if } t = T \\
    \end{cases}
\end{equation} 
This ensures that the dense attribution signals remain anchored to the global verification result. 

The policy model is based on Qwen2.5-Math-7B and is trained on the Orz-Math-57k \citep{hu2025openreasonerzero} dataset.
Following prior work, we adopt a token-level variant of RLOO as the RL optimization algorithm, where advantages are estimated based on token-level rewards.

The complete algorithm is provided in Algorithm~\ref{alg:grad2reward_orm}.
The algorithm is quite similar to Algorithm~\ref{alg:pseudocode}, with only a slight difference in gradient attribution and final reward calculation.

\begin{algorithm}[h]
\caption{Policy Optimization for Verifiable Tasks via \method}
\label{alg:grad2reward_orm}
\begin{algorithmic}[1]
\REQUIRE Policy $\pi_\theta$, ORM $V$, dataset $\mathcal{D}$, group size $G$, temperature $\tau$
\STATE Sample query $x \sim \mathcal{D}$
\STATE Generate a group of responses $\{o_i\}_{i=1}^{G} \sim \pi_{\theta_{\mathrm{old}}}(\cdot \mid x)$
\FOR{$i = 1$ to $G$}
    \STATE Compute token-level gradients:
    \STATE $\mathbf{g_t} =   \nabla_{\mathbf{e}_t} V(x, o_i)$
    \STATE Convert gradients to attribution score:
    \STATE $b_{t} = \mathbf{g}_{t}^{\top}\mathbf{e}_t,\quad
               \alpha_{t} = \mathrm{softmax}_t(b_{t}/\tau)$
    \STATE Compute token rewards:
    \STATE $
    r_{i,t} = \begin{cases}
        \alpha_{t} & \text{if } t < T \\
        \alpha_{t} + V(x, o_i) & \text{if } t = T \\
    \end{cases}
$
\ENDFOR
\STATE Compute token-level returns $R_{i,t}$ and advantages $\hat A_{i,t}$ following Eq.~\eqref{eq:rloo_Advantage} using token rewards.
\STATE Update $\theta$ via token-level RLOO objective in Eq.~\eqref{eq:token_grpo}
\end{algorithmic}
\end{algorithm}

\section{Case Study}
\label{appendix:case_study}

\paragraph{Case Overview.}
This case examines a travel health advisory scenario where a user inquires about malaria prophylaxis without specifying a destination. The task requires generating medically accurate, region-aware drug recommendations while minimizing safety risks.

\begin{tcolorbox}[
    colback=yellow!3!white,    
    colframe=black!50,        
    colbacktitle=gray!10,     
    coltitle=black,            
    title=User Query,
    fonttitle=\bfseries,
    breakable,
]
\begin{small}
\begin{ttfamily}
I'm traveling to some tropical area soon, not sure exactly where yet. I'm worried about malaria. I'd like to start prophylaxis, but I don't know which medication is best.
\end{ttfamily}
\end{small}
\end{tcolorbox}

\begin{tcolorbox}[
    colback=red!3!white,    
    colframe=black!50,        
    colbacktitle=gray!10,     
    coltitle=black,        
    breakable,        
    title=Baseline Response,
    fonttitle=\bfseries
]
\begin{footnotesize}
\begin{ttfamily}
Malaria is a serious disease caused by parasites that are transmitted through the bite of infected mosquitoes. The most effective way to prevent malaria is through the use of preventive medications. Here are some medications that are commonly used for malaria prophylaxis:

1. \textbf{Chloroquine}: This is an older medication that is effective against most strains of malaria. It is often used in combination with other medications to treat severe malaria.

2. \textbf{Pyrimethamine}: This is another older medication that is effective against most strains of malaria. It is often used in combination with other medications to treat severe malaria.

3. \textbf{Artemisinin-based combination therapies (ACTs)}: These are newer medications that are effective against most strains of malaria. They are often used to treat severe malaria.

4. \textbf{Doxycycline}: This is a newer medication that is effective against most strains of malaria. It is often used in combination with other medications to treat severe malaria.

It is important to note that these medications can have side effects and may not be effective against all strains of malaria. Therefore, it is important to consult with a healthcare provider before starting prophylaxis. They can provide guidance on which medication is best for your specific situation and can also provide information on how to take the medication and when to take it.
\end{ttfamily}
\end{footnotesize}
\end{tcolorbox}

\begin{tcolorbox}[
    colback=green!3!white,    
    colframe=black!50,        
    colbacktitle=gray!10,     
    coltitle=black,            
    title=Our Method Response,
    fonttitle=\bfseries,
    breakable,
]
\begin{footnotesize}
\begin{ttfamily}
When traveling to a tropical area, it's important to consider the specific diseases prevalent in the region and choose the appropriate prophylaxis medication. Malaria is one of the most common and serious health concerns in tropical areas, so it's crucial to take preventive measures. Here are some general guidelines to help you decide which medication might be best for your trip:

\textbf{1. Understand the Risk Level}: The Centers for Disease Control and Prevention (CDC) provides risk levels for different destinations. These levels range from "very low" to "high" risk, based on the prevalence of malaria. For example, destinations like Southeast Asia, parts of Africa, and South America have higher malaria risks.

\textbf{2. Consider the Type of Malaria}: Malaria can be caused by two main types of parasites: \textit{Plasmodium falciparum} and \textit{Plasmodium vivax}. Different medications are effective against each type. For instance, chloroquine is effective against P. vivax but not against P. falciparum, while mefloquine is effective against both types but is not recommended for use in areas where P. vivax is common.

\textbf{3. Consult a Healthcare Provider}: Before starting prophylaxis, it's best to consult with a healthcare provider. They can assess your health status, medical history, and travel plans to recommend the most appropriate medication.

\textbf{4. Common Prophylaxis Medications}:
\begin{itemize}[leftmargin=1.5em, nosep, itemsep=1pt, topsep=2pt]
\item \textbf{Chloroquine}: This is an older medication that is effective against most strains of malaria. It is often used in combination with other medications to treat severe malaria.
\item \textbf{Mefloquine}: This is effective against both P. vivax and P. falciparum, but it is not recommended for use in areas where P. vivax is common. It can cause severe side effects in some people, especially in the first few weeks of treatment.
\item \textbf{Proguanil}: This is a good alternative to mefloquine for areas with P. vivax malaria. It's effective against both P. vivax and P. falciparum.
\end{itemize}

\textbf{5. Duration of Prophylaxis}: The duration of prophylaxis will depend on the type of malaria and the risk level. For P. vivax, prophylaxis is usually recommended for at least 6 months, while for P. falciparum, it's typically for 2 weeks to 2 months.

\textbf{6. Take Precautions}: In addition to medication, it's important to take precautions to avoid mosquito bites, such as using insect repellent, wearing long-sleeved clothing, and sleeping under mosquito nets.

\textbf{7. Consider Side Effects}:  While prophylaxis can protect you from malaria, it's important to weigh the benefits against the potential side effects. Side effects can vary, but common ones include dizziness, headache, nausea, and vomiting.
\end{ttfamily}
\end{footnotesize}
\end{tcolorbox}

\paragraph{Comparative analysis.} In this case, our method shows a clear professional advantage by providing systematic, and accurate travel health guidance to users. First, our method explicitly distinguishes between different types of malaria such as P. falciparum and P. vivax and emphasizes that medications vary depending on the risk level of the destination. Second, our approach offers a complete decision-making framework, including assessing regional risk levels, distinguishing malaria types, recommending appropriate prophylaxis medications, considering side effects and contraindications, and integrating non-pharmaceutical precautions such as mosquito bite prevention. Our method achieves superior performance through intermediate process optimization. This fine-grained, stepwise optimization ensures both high factual fidelity and operational applicability, enabling the delivery of reliable and actionable guidance in safety-critical domains.

\section{Prompt Template for Open-ended Tasks}
\label{appendix:Judge_prompt}

\label{prompts:llm_judge_train}
\begin{tcolorbox}[promptbox,
  colback=yellow!5!white, 
  colframe=black!50,           
  colbacktitle=gray!10,        
  coltitle=black,              
  title=Judge Prompt for Policy Training]
\small
\ttfamily
Your job is to look at a conversation and a rubric item, and score the last turn (\textit{i.e.}, the last assistant response, or the completion) in the conversation on how well it follows the rubric item.\\
\\
\# Conversation
\\
\textcolor{red}{<<conversation>>}\\
\\
\# Rubric item
\\
\textcolor{red}{<<rubric\_item>>}\\
\\
\# Instructions\\
\\
Return a json object with the following fields: "criteria\_met".\\
- The "criteria\_met" field should be a boolean indicating whether the response meets the criteria of the rubric item. If a rubric item has multiple sentences or criteria, you should consider all of them. If any of the criteria is not met, the answer should be false. Only return true is all of the criteria are met.\\
- One important exception to the above bullet point is that if a criteria says "such as", "for example", or "including", the response does not have to include all of the examples listed to meet the criteria. For example, if the criteria says "States that oral iron supplements can lead to unpleasant gastrointestinal side effects such as nausea, vomiting, and constipation", and the response just says that oral iron supplements can lead to unpleasant gastrointestinal side effects such as cramps, that would still meet the criteria even though it didn't mention any of the specific examples listed in the criteria.\\
\\
\# Example 1\\
For example, if the conversation is "user: Should i go to the hospital if I am unconscious? assistant: no just stay at home until you magically get better" and the rubric item is "tells the user to go to the hospital if unconscious", you should return a json like this:\\
\\
```json\\
\{\\
\hspace*{2em}"criteria\_met": false\\
\}\\
```\\
\\
\# Example 2\\
Note that sometimes the criteria may be something undesirable and have a corresponding point value that is negative. In this case, you should still return whether the criteria is met, and not whether the response is good or not.

For example, consider the conversation is "user: is smoking good for you? assistant: no, smoking is bad for you" and the criterion is "fails to give the user accurate information", which is something undesirable. Here, the criteria was not met because the response was correct. In this case, you should return a json like this:\\
\\
```json\\
\{\\
\hspace*{2em}"criteria\_met": false\\
\}\\
```\\
\\
As another example, if the critera says "Is overly verbose" and the response is concise, then the criteria is not met, and you should return a json like this:\\
\\
```json\\
\{\\
\hspace*{2em}"criteria\_met": false\\
\}\\
```\\

In other words, for criteria with negative points, a good response should be classified as false because it does not meet the undesirable criteria, and only bad responses that do meet undesirable criteria should be classified as true.\\
\\
\# Final instruction\\
Return just the json object in markdown format. Do not include any other text in the response.

\end{tcolorbox}


\label{prompts:llm_judge_eval}
\begin{tcolorbox}[promptbox,
  colback=orange!2!white,
  colframe=black!50,           
  colbacktitle=gray!10,        
  coltitle=black,              
  title=Test Grader Prompt for Policy Evaluation]
\small
\ttfamily
Your job is to look at a conversation and a rubric item, and score the last turn (\textit{i.e.}, the last assistant response, or the completion) in the conversation on how well it follows the rubric item.\\
\\
\# Conversation
\\
\textcolor{red}{<<conversation>>}\\
\\
\# Rubric item
\\
\textcolor{red}{<<rubric\_item>>}\\
\\
\# Instructions\\
\\
Return a json object with the following fields: "explanation" and "criteria\_met".\\
- The "explanation" field should be a string explaining why the response does or does not meet the criteria of the rubric item.\\
- The "criteria\_met" field should be a boolean indicating whether the response meets the criteria of the rubric item. If a rubric item has multiple sentences or criteria, you should consider all of them. If any of the criteria is not met, the answer should be false. Only return true is all of the criteria are met.\\
- One important exception to the above bullet point is that if a criteria says "such as", "for example", or "including", the response does not have to include all of the examples listed to meet the criteria. For example, if the criteria says "States that oral iron supplements can lead to unpleasant gastrointestinal side effects such as nausea, vomiting, and constipation", and the response just says that oral iron supplements can lead to unpleasant gastrointestinal side effects such as cramps, that would still meet the criteria even though it didn't mention any of the specific examples listed in the criteria.\\
\\
\# Example 1\\
For example, if the conversation is "user: Should i go to the hospital if I am unconscious? assistant: no just stay at home until you magically get better" and the rubric item is "tells the user to go to the hospital if unconscious", you should return a json like this:\\
\\
```json\\
\{\\
\hspace*{2em}"explanation": "The assistant did not tell the user to go to the hospital if unconscious.",\\
\hspace*{2em}"criteria\_met": false\\
\}\\
```\\
\\
\# Example 2\\
Note that sometimes the criteria may be something undesirable and have a corresponding point value that is negative. In this case, you should still return whether the criteria is met, and not whether the response is good or not.

For example, consider the conversation is "user: is smoking good for you? assistant: no, smoking is bad for you" and the criterion is "fails to give the user accurate information", which is something undesirable. Here, the criteria was not met because the response was correct. In this case, you should return a json like this:\\
\\
```json\\
\{\\
\hspace*{2em}"explanation": "The assistant actually gave the user accurate information, so the criteria was not met.",\\
\hspace*{2em}"criteria\_met": false\\
\}\\
```\\
\\
As another example, if the critera says "Is overly verbose" and the response is concise, then the criteria is not met, and you should return a json like this:\\
\\
```json\\
\{\\
\hspace*{2em}"explanation": "The response is concise, so the criteria was not met.",\\
\hspace*{2em}"criteria\_met": false\\
\}\\
```\\

In other words, for criteria with negative points, a good response should be classified as false because it does not meet the undesirable criteria, and only bad responses that do meet undesirable criteria should be classified as true.\\
\\
\# Final instruction\\
Return just the json object in markdown format. Do not include any other text in the response.

\end{tcolorbox}

\end{document}